# Dataset for Robust and Accurate Leading Vehicle Velocity Recognition

Genya Ogawa[1,*]  Toru Saito[1]  Noriyuki Aoi[2]

[1]SUBARU CORPORATION    [2]SIGNATE Inc.

{lastname.firstname}@subaru.co.jp    n.aoi@signate.co.jp

Recognition of the surrounding environment using a camera is an important technology in Advanced Driver-Assistance Systems and Autonomous Driving, and recognition technology is often solved by machine learning approaches such as deep learning in recent years. Machine learning requires datasets for learning and evaluation. To develop robust recognition technology in the real world, in addition to normal driving environment, data in environments that are difficult for cameras such as rainy weather or nighttime are essential. We have constructed a dataset that one can benchmark the technology, targeting the velocity recognition of the leading vehicle. This task is an important one for the Advanced Driver-Assistance Systems and Autonomous Driving. The dataset is available at  https://signate.jp/competitions/657
.

## 1. Introduction

Recognition of the surrounding environment using a camera is an important technology in advanced driver assistance system and autonomous driving, and it has been installed in many vehicles in recent years. We have developed EyeSight using a stereo camera to reduce traffic accidents and driver's burden.

Recognition technology is often developed using machine learning approaches such as deep learning, which has undergone remarkable technological evolution in recent years. In machine learning, it is necessary to prepare a dataset that includes appropriate input and output of the recognition task for training and evaluation, but in order to develop robust recognition technology in the real world, in addition to the normal driving environment, it is important to include data of difficult scenes for the camera, such as at night or in the rain.

Recently, many attempts have been made to promote technological advancement through open innovation, which publishes datasets so that take benchmarks of developed technologies and establishes a framework in which developers in the world can compete their scores of their technology.

In order to carry out the above efforts, we constructed a dataset closed to practical use for development of technology. There are various recognition targets in in-vehicle camera image recognition, but our target task is recognizing the velocity of the leading vehicle [1]. Velocity recognition of the leading vehicle is an important capability for realizing automatic braking to avoid a rear-end collision and adaptive cruise control that maintains a certain distance. At the same time, the task requires many technical elements such as detection of the leading vehicle, tracking objects between frames, recognition of vehicle position in 3D. In the dataset, in addition to the target vehicle velocity, we prepared the vehicle position in the image and the distance to the leading vehicle. We expect this dataset will be used for the development of individual technical elements such as detection or tracking.

We also report the results of a competition held to confirm the usefulness of the dataset.

## 2. Related Work

Currently, many datasets are open to the public. ImageNet[2], which is used for many machine learning image recognitions, not limited to in-vehicle camera images, is a very famous data set, and is also used for the pre-training of deep neural network. In-vehicle camera images, such as MS COCO[3], Cityscapes[4], Mapilally[5], BDD100K[6] are used for detection of recognition targets and semantic segmentation. However, most of these data are accumulated from normal driving environments. There are few data from adverse environments such as nighttime or rainy weather in the datasets. Regarding the adverse environment, some datasets such as DAWN[7] are also open to the public, but its images are collected from the internet. Although conceptual learning and evaluation of adverse environments can be performed, it is difficult to use for developing recognition technology for fixed camera systems.

Furthermore, KITTI[8] is famous as a dataset used for benchmarking. KITTI contains various tasks, but it is a set of different technological tasks such as object detection, 3D recognition, tracking, etc. It is difficult to use it directly for the development of vehicle functions such as autonomous emergency braking (AEB) or adaptive cruise control (ACC).

The dataset we built has the following features.
- Images are taken by the camera of our commercial off-the-shelf system (EyeSight)
- Including many adverse environment scenes such as nighttime or rainy weather
- Sufficient amount of data to be used for machine learning
- Easy to use for developing vehicle functions

In particular, the first point is valuable because this type of dataset is rarely open to public. Therefore, this dataset is more useful for developing technology to a more practical level than other datasets.

### 3. Dataset

#### 3.1 Overview

The constructed dataset consists of images from a stereo camera, disparity image generated from the left and right images from the stereo camera, vehicle information for each frame (ego vehicle velocity, steering angle, parameter to covert distance from disparity), position (rectangular) information on the image of the leading vehicle, distance to the leading vehicle and velocity of the leading vehicle.

#### 3.2 Images

Regarding the stereo camera image, a right camera image is 1000 x 420 pix, a left camera image is 1128 x 420 pix, and the frame rate is 10 fps (Fig.1). One scene contains 100 to 200 frames. The dataset includes 1000 scenes that ego vehicle is following the leading vehicle. The scenes in which leading vehicle were within 120m were selected.

In the dataset, mp4 compression was applied to reduce the data size. We confirmed that the compression did not significantly affect the recognition performance of a general detection model. For releasing the dataset publicly, the vehicle license plates and pedestrian faces which may identify personal information are blurred in each area. We checked that there was no significant effect on the recognition performance after blurring. To keep personal identifiable information, we also confirmed that there were no readable nameplates for house in the dataset.

For making disparity image, the right camera image is divided into areas of 4 × 4 pixel block, and the 1/16 pixel level resolution disparity is produced for each area by area-based matching (Fig.2). If an area contains little image features, the disparity is set to 0 because it is not reliable. Since the disparity image is information related to get accurate distance to the leading vehicle, we do not compress it.

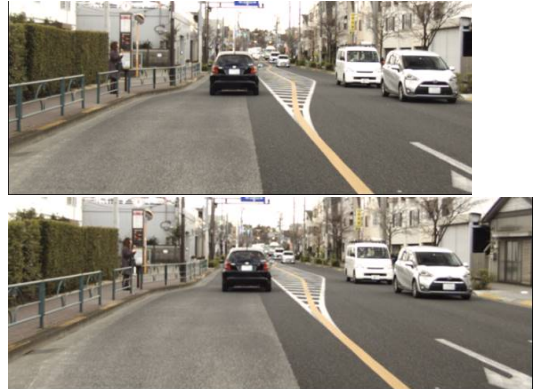

Fig.1 right and left images

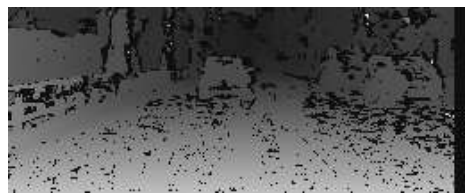

Fig.2 disparity image (correction is performed for viewing)

#### 3.3 Leading vehicle position information on the image

As shown in Fig. 3, the position on the image of the leading vehicle is rectangular information circumscribing the preceding vehicle in the right camera image. In the training data, each frame has the position information, but in the test data, only the first frame has that information. The information is provided at the first frame in order to specify which vehicle in the image is the leading vehicle.

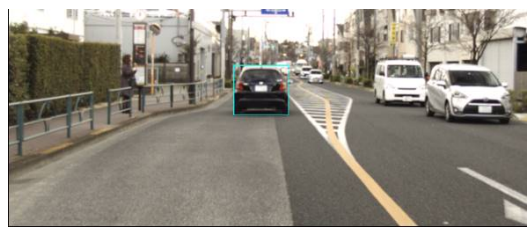

Fig.3   position information of leading vehicle

#### 3.4 Distance and velocity of leading vehicle

The distance and velocity of the leading vehicle were collected using IBEO's laser radar. Since the distance measurement accuracy is high, the velocity is calculated by differentiating the distance and averaging 7 frames before and after the frame. The relative speed can be obtained by differentiating the distance, so the absolute speed value is obtained by adding the ego vehicle velocity to the relative velocity (Fig.4). Since the leading vehicle velocity is generated using future frames, to recognize the velocity partially requires future prediction. Difficult scenes to detect the leading vehicle by the laser radar are not included in the dataset.

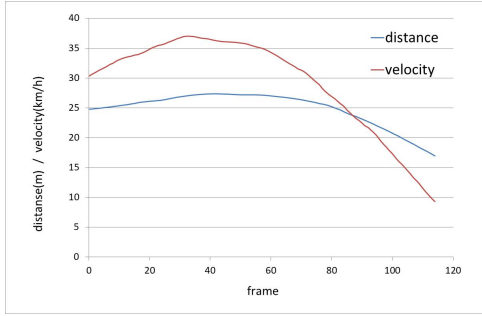

Fig.4 distance and velocity of leading vehicle

### 3.5 Scene

Fig.5-8 show the frequency distribution of driving scenes. We considered the distribution of general drivers' driving scenes, but we included higher number of scenes in nighttime or rainy weather environment.

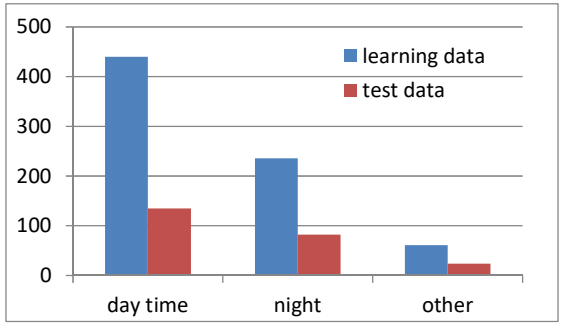

Fig.5 time zone

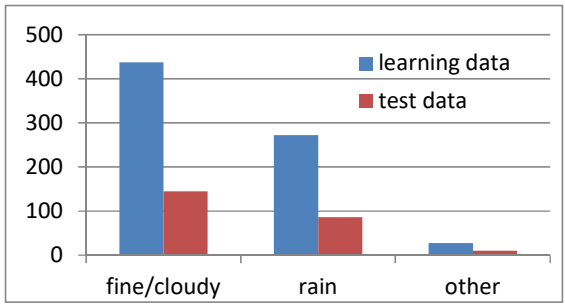

Fig.6 weather

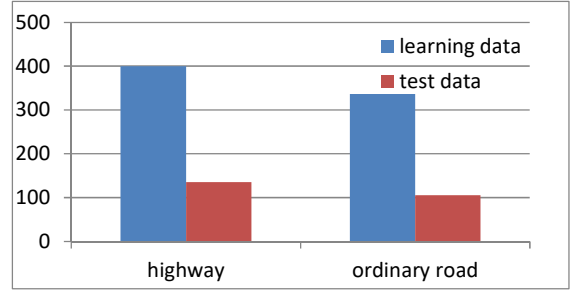

Fig.7 driving place

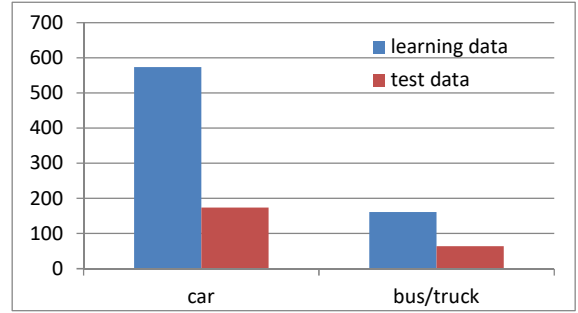

Fig.8 type of leading vehicle

### 4. Evaluation metric for benchmark

Evaluation metrics were defined such that benchmark evaluation could be performed using this dataset. The evaluation value is calculated by Eq. (1) for each scene. $n_j$ is the number of frames in scene j, $p_i$ and $t_i$ are the prediction value and target value at frame i of scene j respectively. The first 19 frames are not evaluated because the recognition task is a time series problem. In addition, the error between the prediction value and the target value is the basis of the evaluation, but in order to prevent the influence of the scenes with the high vehicle velocity, we adjust the metric by the limitation and applying the gain according to the velocity.

$$e_j = \frac{1}{n_j - 19} \sum_{i=20}^{n_j} \min\left(\left|\frac{p_i - t_i}{0.07 t_i + 3}\right|, 1\right) \quad (1)$$

The final evaluation metric is obtained by taking the weighted average of each scene by Eq. (2). $w_j$ is the weight of scene j.

In the scene where the distance to the leading vehicle is decreasing, the weight is set to 3, and the weight is 1 otherwise. The scenes where the distance is decreasing are important scenes for applications because the risk of collision to the leading vehicle increase.

$$error = \frac{\sum_{j=1}^{s} w_j e_j}{\sum_{j=1}^{s} w_j} \qquad (2)$$

We defined the above evaluation metrics, however, following capability to the leading vehicle and stability are also important factors for the applications. Therefore, we investigated the relationships between the evaluation value, following capability (recognition delay), and stability.

The measurement of the following capability is obtained by assuming that the target value shifted by N frames from the correct target data as the recognition output value. The results are shown in Table.1. Since the evaluation values are distributed in 0-1, 10 frames (1 second) recognition delay highly deteriorates the result. Therefore, we think that the following capability can be measured by this evaluation metric. Moreover, if the correct target data is constant (the leading vehicle is at a constant relative velocity), the evaluation value should not be deteriorated in this setting. Thus, we can interpret that this dataset includes many scenes that the relative velocity to the leading vehicle largely changed.

We also evaluate the stability as follows. For each frame, we use 2A-B as prediction value to evaluate the stability where A is the target value for the frame and B is the average of total 2M-1 frames before and after the frame. The results are shown in Table.2. These results show that the stability does not affect to the evaluation metrics rather than the following capability. Therefore, this evaluation metric relatively emphasizes the following capability than the stability. We suppose, however, this tendency is because of the smoothness of the original target values. The actual recognition results have some error. In equation (1), the evaluation value is about 0.1 by an error of around 0.3 km/h if the vehicle speed is 0 km/h. Also, about 0.7 km/h error leads to 0.1 evaluation value if the vehicle speed is 100 km/h. Therefore, we conclude that the stability is considered to some extent in this evaluation metric. If the behavior of the actual leading vehicle is unstable, it leads to lowering the evaluation value, so it is essential to take the stability into consideration to control the vehicle in each application.

Table.1 following capability

| N | 0 | 5 | 10 | 15 | 20 |
|---|---|---|----|----|----|
| error | 0.000 | 0.097 | 0.174 | 0.231 | 0.271 |

Table.2 stability

| M | 1 | 5 | 9 |
|---|---|---|---|
| error | 0.003 | 0.013 | 0.025 |

5. Data set usefulness verification

In order to verify the usefulness of this dataset as a benchmark, we held a competition to compete for the evaluation metrics defined in Chapter 4.

The competition was held on the platform operated by SIGNATE Co., Ltd (https://signate.jp/). The timeline was two and a half months. Any individuals, organizations, nationalities, etc. could participate. As restrictions, memory must be 32 GB or less (GPU memory must be 10 GB or less), model file size limit was 2GB or less.

As a result, there were 126 teams participated and 2077 submissions. The results of the top 3 teams are shown in table.3. To use the prediction results for the application, we require that the evaluation value should be under 0.1. The top 3 team's results exceeded that requirement.

Table3. evaluation value at competition

| 1st | 0.0846 |
|-----|--------|
| 2nd | 0.0915 |
| 3rd | 0.0917 |

The overall trend was that submissions from all teams had similar tendency of scenes in which they received lower scores. Figure 9 shows images that the average evaluation values of top 20 teams that were below 0.3.

Since there are many scenes of windshield water droplets and wipers shielding at night or in the rain, which are difficult scenes in our actual development. Therefore, we believe this dataset is an appropriate one for the leading vehicle recognition task by the camera.

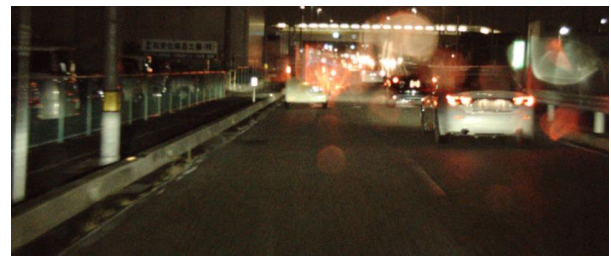

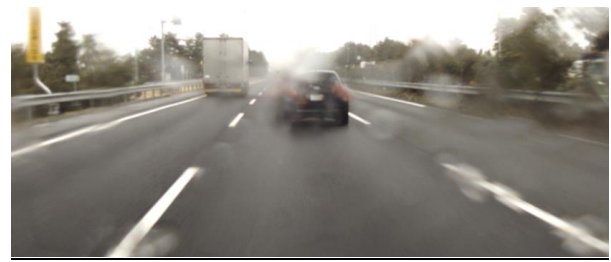

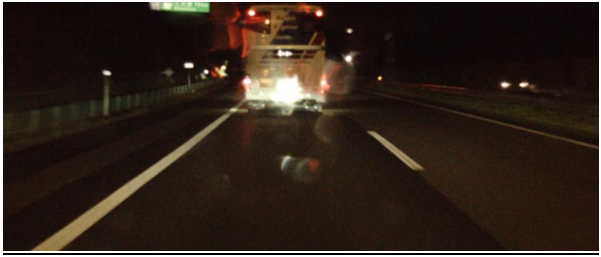

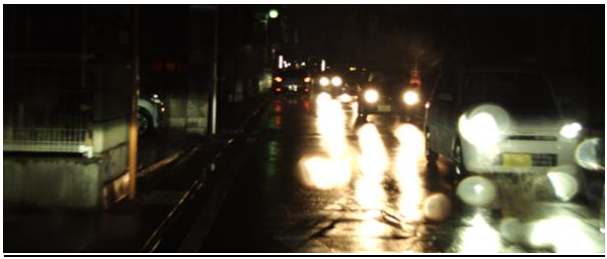

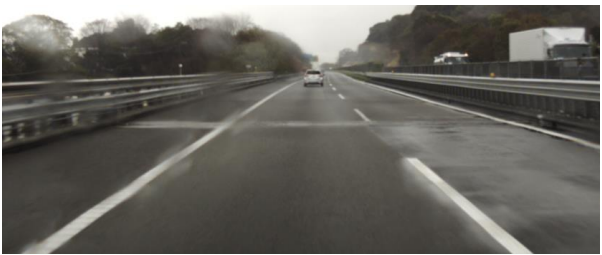

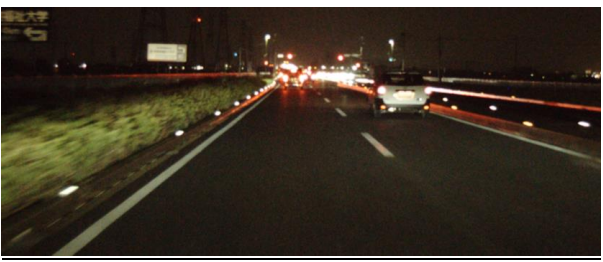

Fig.9 Scenes with low evaluation value

6. Conclusion

We construct a dataset for velocity recognition of the leading vehicle, which is an important task in advanced driver assistance system and autonomous driving. The dataset contains camera images from our product system. The dataset contains many adverse environmental scenes such as nighttime or rainy weather. It can be used as a benchmark for recognition technology.

The dataset is available at **https://signate.jp/competitions/657**, and it is possible to post the prediction result and measure the performance of the prediction. We are eagerly awaiting the future development of various tasks by utilizing this dataset.

Acknowledgments

We would like to thank all the participants of the competition.

References

(1) T.Saito, T.Okubo, and N.Takahashi, "Robust and Accurate Object Velocity Detection by Stereo Camera for Autonomous Driving," in Proc. IEEE Intelligent Vehicle Symposium', 2020.

(2) DENG J. "A Large-Scale Hierarchical Image Database" IEEE Computer Vision and Pattern Recognition (CVPR), 2009, pp. 248-155.

(3) T.Lin, M.Maire, S.Belongie, J.Hays, P.Perona, D.Ramanan, P.Doll´ar, and C.Zitnick, "Microsoft COCO: Common objects in context," in European Conference on Computer Vision, 2014, pp. 740–755.

(4) M.Cordts, M.Omran, S.Ramos, T.Rehfeld, M.Enzweiler, R.Benenson, U.Franke, S.Roth, and B.Schiele, "The cityscapes dataset for semantic urban scene understanding," in Proc. of the IEEE Conference on Computer Vision and Pattern Recognition (CVPR), 2016.

(5) G.Neuhold, T.Ollmann, S.Bulo, and P.Kontschieder, "The mapillary vistas dataset for semantic understanding of street scenes," in Proceedings of the IEEE International Conference on Computer Vision, 2017, pp. 4990–4999.

(6) F.Yu, W.Xian, Y.Chen, F.Liu, M.Liao, V.Madhavan, and T.Darrell, "Bdd100k: A diverse driving video database with scalable annotation tooling," arXiv:1805.04687, 2018.

(7) M.Kenk, and M.Hassaballah. "DAWN: vehicle detection in adverse weather nature dataset." arXiv preprint arXiv:2008.05402 (2020).

(8) A.Geiger, P.Lenz, and R.Urtasun, "Are we ready for autonomous driving? the kitti vision benchmark suite," in IEEE Conference on Computer Vision and Pattern Recognition, 2012, pp. 3354–3361.